\documentclass[letterpaper]{article} % DO NOT CHANGE THIS
\usepackage{aaai2026} %\usepackage[submission]{aaai2026}  % DO NOT CHANGE THIS
\usepackage{times}  % DO NOT CHANGE THIS
\usepackage{helvet}  % DO NOT CHANGE THIS
\usepackage{courier}  % DO NOT CHANGE THIS
\usepackage[hyphens]{url}  % DO NOT CHANGE THIS
\usepackage{graphicx} % DO NOT CHANGE THIS
\usepackage{natbib}  % DO NOT CHANGE THIS AND DO NOT ADD ANY OPTIONS TO IT
\usepackage{caption} % DO NOT CHANGE THIS AND DO NOT ADD ANY OPTIONS TO IT
\usepackage{algorithm}
\usepackage{algorithmic}
\usepackage{amssymb}
\usepackage{booktabs}
\usepackage{multirow}
\usepackage{subcaption}
\usepackage{breqn}
\usepackage{pifont}
\usepackage{colortbl}
\newcommand{\cmark}{\ding{51}}%
\newcommand{\xmark}{\ding{55}}%

\usepackage{newfloat}
\usepackage{listings}
\DeclareCaptionStyle{ruled}{labelfont=normalfont,labelsep=colon,strut=off} % DO NOT CHANGE THIS
\floatstyle{ruled}
\newfloat{listing}{tb}{lst}{}
\floatname{listing}{Listing}
\nocopyright % \nocopyright -- Your paper will not be published if you use this command
\title{STADE: Standard Deviation as a Pruning Metric}
\author {
    Diego Coello de Portugal Mecke\textsuperscript{\rm 1,\rm 2},
    Haya Alyoussef\textsuperscript{\rm 1},
    Maximilian Stubbemann,
    Ilia Koloiarov\textsuperscript{\rm 1,\rm 2},
    Tom Hanika\textsuperscript{\rm 1},
    Lars Schmidt-Thieme\textsuperscript{\rm 1,\rm 2}
}
\affiliations {
    \textsuperscript{\rm 1} ISMLL, University of Hildesheim\\
    \textsuperscript{\rm 2} VWFS DARC
}

\begin{document}

\maketitle

\begin{abstract}
Recently, Large Language Models (LLMs) have become very widespread and are used to solve a wide variety of tasks.
To successfully handle many of these tasks, LLMs require longer training times and larger model sizes.
This makes LLMs ideal candidates for pruning methods that reduce computational demands while maintaining performance.
Previous methods require a retraining phase after pruning to maintain the original model's performance.
However, state-of-the-art pruning methods, such as Wanda, prune the model without retraining, making the pruning process faster and more efficient.
Building upon Wanda’s work, this study provides a theoretical explanation of why the method is effective and leverages these insights to enhance the pruning process.
Specifically, a theoretical analysis of the pruning problem reveals a common scenario in Machine Learning where Wanda is the optimal pruning method.
Furthermore, this analysis reveals cases where Wanda is no longer optimal. To tackle those cases, we develop a new method, \textit{STADE}, based on the standard deviation of the input.
From a theoretical and empirical standpoint, \textit{STADE} demonstrates better generality across different scenarios.
Finally, extensive experiments on Qwen, Llama and Open Pre-trained Transformers (OPT) models validate these theoretical findings, showing that depending on the training conditions, Wanda's optimal performance varies as predicted by the theoretical framework.
These insights contribute to a more robust understanding of pruning strategies and their practical implications.
% Code is available at: https://github.com/Coello-dev/STADE/
\end{abstract}

\begin{links}
    \link{Code}{https://github.com/Coello-dev/STADE/tree/main}
    % \link{Datasets}{https://aaai.org/example/datasets}
    % \link{Extended version}{https://aaai.org/example/extended-version}
\end{links}

\section{Introduction}

Large Language Models (LLMs) \cite{gpt1, gpt2, gpt3} have revolutionized not only the field of Natural Language Processing (NLP) but also numerous real-world applications that affect everyday life. Their ability to generate coherent text, perform complex reasoning, and support a variety of conversational and decision-making tasks has led to widespread adoption in both research and industry. With the advent of increasingly autonomous systems \cite{agent1, agent2, agent3}, these models now assist with tasks ranging from content creation and translation to automated customer support and strategic decision making.

Despite these impressive capabilities, LLMs are notorious for their substantial computational requirements \cite{kaplan2020scalinglawsneurallanguage}. The high memory footprint, extensive processing power, and significant energy consumption often limits their deployment on devices with limited resources, such as mobile phones or embedded edge devices. In addition, the large-scale training of these models contributes to increased operational costs and a non-negligible environmental impact. Consequently, the drive to reduce the computational and storage demands of LLMs has become a central focus in the field \cite{llmcomputation}.

To mitigate these computational challenges, a variety of approaches have been explored. One prominent strategy involves reducing the storage requirements of model weights through \textit{quantization} \cite{1bit, quantization}. Quantization techniques lower the numerical precision of weights and activations, resulting in reduced memory usage and accelerated inference speeds, often with minimal degradation in performance. Another effective approach is to remove unimportant weight parameters through \textit{pruning} \cite{obd}. Pruning methods seek to eliminate redundancies in the network by removing weights that contribute little to overall model performance, thereby decreasing both the computational load and the inference latency.

Pruning techniques can be applied during training \cite{pruning_training} or after the model has been fully trained, in what is known as \textit{post-training pruning} \cite{slicegpt}. The latter approach is particularly appealing when the goal is to adapt a pre-trained model for deployment on resource-constrained devices, as the main challenge is not the training process but rather fitting the model into a limited hardware environment. Although some post-training pruning strategies involve costly retraining steps \cite{softprune, xubesa}, previous studies \cite{wanda, sparsegpt} have demonstrated that a model can maintain a large fraction of its original performance even when up to 50\% of its weights are pruned without any retraining.

A notable pruning method is Wanda \cite{wanda}, which employs a simple yet effective strategy based on the $L_2$-$loss$ to guide weight removal. Despite its empirical success, the fundamental reason for the superior performance of the $L_2$-$loss$ over alternative norms (e.g., $L_1$ or $L_{\infty}$) was neither formally analyzed or fully understood. As noted in the original paper:
\textit{"We find that }$l_2$\textit{ norm tends to work better than other norm functions (e.g., }$l_1$\textit{ and }$l_{\infty}$\textit{) in measuring activation magnitudes. This is possibly because }$l_2$\textit{ norm is generally a smoother metric"} \cite{wanda}.
Such observations have motivated deeper theoretical investigations into pruning criteria.

This work aims to provide a comprehensive analysis of the pruning problem. The contributions are as follows:

\begin{itemize}
	\item A theoretical analysis of the pruning problem is presented, revealing a characterization of machine learning scenarios where \textit{Wanda} emerges as the optimal pruning method.
	\item The analysis is extended to cases where \textit{Wanda}'s approach is suboptimal, thereby motivating the development of a new method, \textit{STADE}.
    \item Multiple experiments with different LLM model families validate empirically the theoretical analysis.
	\item Additionally, an ablation of layer-specific characteristics demonstrates that different pruning metrics can yield better performance when applied selectively across different layers of a model. To the best of our current knowledge, this is the first work to apply distinct pruning metrics to different layers, resulting in improved overall pruning effectiveness.
\end{itemize}

Extensive experiments have been performed across multiple models and configurations to validate the theoretical insights and assess the performance of the proposed \textit{STADE} method. The experiments evaluate perplexity, on different pruning metrics and on different layers for various models, and reveal that the impact of pruning is highly dependent on the statistical properties of the input at each layer. These findings offer valuable guidance for future research in model compression and the efficient deployment of LLMs on consumer devices.

\section{Related Work}

The study of sparse subnetworks within large neural networks has been an area of intense investigation, particularly following the introduction of the \textit{Lottery Ticket Hypothesis} \cite{frankle2019lotterytickethypothesisfinding}. This hypothesis proposes that within a randomly initialized neural network there exist subnetworks (or "winning tickets") that, when trained in isolation, can achieve performance on par with the full network. Subsequent investigations \cite{morcos2019ticketwinallgeneralizing, frankle2020linearmodeconnectivitylottery} have further elucidated the generalization capabilities and connectivity properties of these subnetworks, providing a theoretical basis for pruning methods.

Pruning strategies have evolved significantly over the past decade. Early methods relied on simple heuristics such as magnitude-based pruning \cite{magnitudeprune}, which removes weights with the smallest absolute values under the assumption that these contribute least to network performance. This basic approach laid the groundwork for more sophisticated techniques that consider additional information about the network. For instance, the work in \cite{han2015learningweightsconnectionsefficient} utilized the $L_2$ norm to evaluate the importance of weights, demonstrating that many redundant parameters could be pruned without significant loss in accuracy.

Advancements in pruning have also led to the development of methods that incorporate second-order information. The Optimal Brain Surgeon (OBS) algorithm \cite{optimalbs}, for example, leverages the Hessian matrix of the loss function to estimate the impact of removing individual weights. Although OBS provides more refined pruning decisions, its high computational complexity has restricted its practical application in large-scale models.

More recent approaches have shifted focus to dynamic pruning strategies that are integrated into the training process \cite{pruning_training, chen2021chasing}. These methods progressively reduce the number of active parameters during training, often resulting in models that are sparser and more computationally efficient. However, such strategies may conflict with the scaling laws observed for LLMs \cite{llmscalinglaws}, where performance improvements are closely tied to increases in model size, computational resources, and data availability. As a consequence, post-training pruning techniques have emerged as a pragmatic solution for adapting large pre-trained models to resource-limited environments.

A wide range of post-training pruning techniques has been proposed in recent years. Some methods, such as LoRA-based pruning \cite{prunelora}, incorporate low-rank adaptations to guide the pruning process.  However, retraining the pruned model often incurs significant computational overhead. Others, like SparseGPT \cite{sparsegpt}, use Hessian-based metrics to carefully select which weights to remove, and adjusting the remaining parameters accordingly, thereby preserving critical network functionality. Additionally, strategies that minimize local reconstruction errors within individual blocks \cite{softprune, sparsellm} or layers \cite{layerprune} of Transformer-based architectures have been investigated, underscoring the notion that different layers may require tailored pruning criteria. Some layer-wise pruning techniques employ structured sparsity, assigning a learned importance weight to each matrix, thereby determining its optimal sparsity level \cite{li2024discovering}. Others adopt a block-wise grouping strategy, optimizing sets of layers collectively \cite{xubesa} to balance sparsity and accuracy.

A central aspect of all pruning methodologies is the selection of an appropriate pruning metric that accurately distinguishes between essential and redundant weights. The metric adopted in Wanda \cite{wanda}—which involves computing the \(L_2\) norm of the input and multiplying it by the absolute value of the corresponding weight—has garnered considerable attention for its simplicity and effectiveness. This approach provides a smooth, continuous measure that captures the contribution of each weight to the overall activations. In contrast, more elaborate metrics, including those based on second-order derivatives or layer-specific statistical properties, may offer theoretical advantages but often come at the cost of increased computational overhead.

Overall, the evolution of pruning methods reflects a broader trend in machine learning towards achieving a balance between model efficiency and predictive performance. Early heuristic methods have given way to more principled approaches that take into account the underlying statistics and structure of the network. The continued development of these techniques is critical for the deployment of large-scale neural networks on platforms with limited computational resources. The insights provided by previous studies serve as a valuable foundation for the enhancements presented in this work, including the development of the \textit{STADE} method, which refines pruning strategies by incorporating the statistical characteristics of layer inputs.

\section{Methodology}\label{sec:math_derivation}

Consider a data matrix $X \in \mathbb{R}^{N \times M}$ and a weight matrix $\mathbb{W} \in \mathbb{R}^{M \times H}$, where $N$ is the number of instances in the dataset, $M$ represents the number of features and $H$ represents the number of output features. In Wanda \cite{wanda}, the pruning of each column $\mathbb{W}_{:,i} \in \mathbb{R}^{M}$ is performed according to the criterion:

\begin{equation}
    \min_j \|X_{:,j}\|_2 |\mathbb{W}_{j,i}|
\end{equation}

where $\|X_{:,j}\|_2$ is the $L_2$-$norm$ of feature $j$ in the dataset.
In the following section, the pruning problem is formalized and it is demonstrated that \textit{Wanda} selection criterion is optimal for layers with a centered inputs, i.e., inputs whose expected value in each coordinate is $0$. With this insight, a generalization to layers with uncentered inputs is derived, leading to the proposed method \textit{STADE}.

\subsection{Problem definition}

Let $X \in \mathbb{R}^M$ be a random multivariate variable with $\mu_i = \mathbb{E}[X_i]$ and $\sigma_i=Var[X_i]$, and consider a linear layer with a weight matrix $\mathbb{W} \in \mathbb{R}^{M \times H}$ and a bias term $\mathbb{B} \in \mathbb{R}^{H}$. The pruning process for the \(m\)-th column will be focused on the weight matrix (denoted by $W = \mathbb{W}_{:,m} \in \mathbb{R}^M$) and the corresponding bias (denoted by $B = \mathbb{B}_{m} \in \mathbb{R}$). In this setting, the pruning problem aims to find the optimal $W^* \in \mathbb{R}^M$ and $B^* \in \mathbb{R}$  such that:

\begin{equation}\label{eq:problem_definition}
    \begin{split}
        \min_{W^*,B^*} \mathbb{E}[\left( (B + \sum_{i=1}^M X_i W_i) - (B^* + \sum_{i=1}^M X_i W^*_i)\right)^2]\\
        \text{ s.t. } \forall i \in \{1,...,M\} \backslash \{j\}, W^*_i = W_i, W_j^*=0
    \end{split}
\end{equation}

Note that the objective is to select the pruning weight \(W_j\) so that the output remains almost unchanged, while allowing the bias term to be updated.

\subsection{STADE derivation}

Starting from the formulation in Eq. \ref{eq:problem_definition}, the objective function can be reformulated as follows:

\begin{equation}\label{eq:first_derivation}
    \begin{split}
        \mathbb{E} & [\left( (B +\sum_{i=1}^M X_i W_i) - (B^* + \sum_{i=1}^M X_i W^*_i) \right)^2]\\
        =& \mathbb{E}[\left( (B - B^*) + X_j W_j\right)^2]\\
        =& \mathbb{E}[(B - B^*)^2 + 2(B - B^*)(X_j W_j) + (X_j W_j)^2\\ 
        =& (B - B^*)^2 + 2(B - B^*)\mathbb{E}[X_j W_j] + \mathbb{E}[(X_j W_j)^2]\\
        =& (B - B^*)^2 + 2(B - B^*) \mu_j W_j + (\sigma_j^2 + \mu_j^2)W_j^2
    \end{split}
\end{equation}

To determine the optimal solution of the convex problem (with respect to $B^*$) in Eq. \ref{eq:first_derivation}, the derivative is computed to obtain the stationary and minimum point:

\begin{equation}\label{eq:optimal_bias}
	\begin{split}
    	&\frac{d}{d B^*} [\mathbb{E} [\left( (B + \sum_{i=1}^M X_i W_i) - (B^* + \sum_{i=1}^M X_i W^*_i)\right)^2]]\\
    	&= \frac{d}{d B^*}[(B - B^*)^2 + 2(B - B^*) \mu_j W_j + (\sigma_j^2 + \mu_j^2)W_j^2]\\
    	&= -2(B - B^*) - 2\mu_j W_j = 0 \Leftrightarrow B^* = \mu_j W_j + B
	\end{split}
\end{equation}

Substituting the optimal bias in Eq. \ref{eq:first_derivation} yields the solution for $W^*$:

\begin{equation}\label{eq:optimal_wj}
	\begin{split}
     	&\min_{W^*,B^*} \mathbb{E}[\left( (B + \sum_{i=1}^M X_i W_i) - (B^* + \sum_{i=1}^M X_i W^*_i) \right)^2] \\
     	&=\min_{j,B^*} (B - B^*)^2 + 2(B - B^*) \mu_j W_j + (\sigma_j^2 + \mu_j^2)W_j^2\\
        &= \min_{j} (B - (\mu_j W_j + B))^2 + 2(B - (\mu_j W_j + B)) \mu_j W_j \\
        &+ (\sigma_j^2 + \mu_j^2)W_j^2  = \min_{j} (\mu_j W_j)^2 - 2(\mu_j W_j) \mu_j W_j\\
        &+ (\sigma_j^2 + \mu_j^2)W_j^2 = \min_{j} \sigma_j^2 W_j^2  \approx \min_{j} \frac{||X_{:,j} - \mu_j||_2^2}{N-1} W_j^2\\
        &\approx \frac{1}{N-1} \left( \min_{j} ||X_{:,j} - \frac{1}{N}\sum_{n=1}^N X_{n,j}||_2 |W_j| \right)^2
	\end{split}
\end{equation}

Since the our goal is to find the optimal $j$ that minimizes the loss ($arg\min_j$), the factor $\frac{1}{N-1}$ and the squaring operation can be omitted.

\begin{table*}[h]
	\centering
	\begin{tabular}{cccc}
    	\toprule
    	\multirow{2}*{Method} & Weight & Centered & \multirow{2}*{Pruning criterion} \\
    	 & Update & Input &  \\
    	\midrule
    	Magnitude \cite{magnitudeprune} & \xmark & Any & $|W_{i,j}|$ \\[1mm]
    	Wanda \cite{wanda} & \xmark & Any & $||X_{:,j}||_2 |W_{i,j}|$ \\[1mm]
    	Sparsegpt \cite{sparsegpt} & \cmark & Any & $[|W|^2/diag[(X^TX + \lambda \mathbf{1})^{-1}]]_{i,j}$ \\
    	\midrule
    	STADE & \xmark & Any & $||X_{:,j} - \frac{1}{N}\sum_{n=1}^N X_{n,j}||_2 |W_{i,j}|$ \\[1mm]
    	\midrule
    	\multirow{2}*{STADE-W} & \xmark & Yes & $||X_{:,j}||_2 |W_{i,j}|$ \\[1mm]
      	& \xmark & No & $||X_{:,j} - \frac{1}{N}\sum_{n=1}^N X_{n,j}||_2 |W_{i,j}|$ \\[1mm]
    	\bottomrule
	\end{tabular}
	\caption{Comparison of pruning weight metrics across different methods. The column \emph{Centered Input} indicates whether the pruning method distinguishes between inputs with zero mean (Yes), without zero mean (No), or treats them equivalently (Any).}
	\label{tab:pruning_methods_definition}
\end{table*}

\subsection{Wanda derivation}

Many modern Transformers \cite{llama1,llama2,llama3} employ normalization layers. These design choices simplify the original problem by enforcing that the input $X$ is normalized ($\mu_i=\mathbb{E}[X_i]=0$). Incorporating these conditions to the previous derivations (Eqs. \ref{eq:optimal_bias} and \ref{eq:optimal_wj}) leads to:

\begin{equation}\label{eq:wanda_derivation}
	\begin{split}
    	& B^* = \mu_j W_j + B = 0*W_j + B = B\\
        \\
        \min_{j} & ||X_{:,j} - \mu_j||_2 |W_j| = \min_{j} ||X_{:,j}||_2 |W_j|
	\end{split}
\end{equation}

This derivation results in the \textit{Wanda} criterion, where the bias term doesn't need to get updated. Please notice that \textit{Wanda} is optimal under the previous assumptions, i.e., it is only optimal for layers with centered inputs.

\subsection{STADE-W: Using different metrics for different layers}\label{sec:stade_w_theory}

Based on the previous theoretical insight we introduce  \textit{STADE-W}, a pruning strategy that employs different pruning criterions depending on whether the input is normalized. The pruning metrics derived from the previous analysis are as follows:

\begin{alignat}{2}
& \text{Wanda criterion:} \quad ||X_{:,j}||_2 |W_{i,j}| \label{eq:wanda_metric}\\
& \text{STADE criterion:} \quad ||X_{:,j} - \frac{1}{N}\sum_{n=1}^N X_{n,j}||_2 |W_{i,j}| \label{eq:std_bias_metric}
\end{alignat}

\textit{STADE-W} applies the \textit{STADE} metric for biased inputs (such as the second layer of an MLP or the output layer in multi-head attention) and the \textit{Wanda} metric for unbiased inputs (such as the first layer of an MLP or the queries, keys, and values in multi-head attention).
In theory, \textit{STADE} should be able to identify that the mean is 0 and return the same output as \textit{Wanda}. However, in practice the dataset used for calibration might lead to a slightly different mean estimation and therefore, \textit{STADE} ends up underperforming.

\subsection{Optimal pruning metric}

In order to clarify which pruning metric to use in which linear layer we make the following distinctions:

\begin{itemize}
    \item \textbf{Centered inputs}: When the input distribution is centered the optimal method is \textit{Wanda}. This would be the case when the previous layer is a \textit{Batchnorm} \cite{batchnorm},  \textit{Groupnorm} \cite{groupnorm} or \textit{Layernorm} layer \cite{layernorm}, but not in the case of a \textit{RMSnorm} layer \cite{rmsnorm}.
    \item \textbf{Uncentered inputs}: In this case, the input mean is no longer 0 and therefore \textit{Wanda} is no longer optimal. Therefore \textit{STADE} should be use since it takes into account the non-zero mean.
\end{itemize}

This protocol will be used throughout the paper unless specified otherwise. Notice that within the same model, different layers could belong to different scenarios as mentioned before with \textit{STADE-W}.

\begin{table*}[h] 
 \centering 
 \begin{tabular}{cccccccccccccc} 
 \toprule 
 \multirow{2}*{Methods} & \multirow{2}*{Sparsity} & Llama-1 & \multicolumn{2}{c}{Llama-2} & \multicolumn{2}{c}{Llama-3} & \multicolumn{5}{c}{Qwen3} \\
\cmidrule(l{2pt}r{2pt}){3-3} \cmidrule(l{2pt}r{2pt}){4-5} \cmidrule(l{2pt}r{2pt}){6-7} \cmidrule(l{2pt}r{2pt}){8-12}
  & & 7B & 7B & 13B & 3.0-8B & 3.1-8B & 1.7B & 4B & 8B & 14B & 32B \\ 
 \midrule 
 Dense & 0\% & 5.68 & 5.47 & 4.88 & 6.14 & 6.24 & 16.67 & 13.64 & 9.72 & 8.64 & 7.6\\ 
 \midrule 
 Magnitude & 2:4 & 42.53 & 37.76 & 8.89 & 2401.18 & 792.83 & 1808.24 & 1970.45 & 294.48 & 38.58 & 29.89  \\ 
 Wanda & 2:4 & 11.52 & 12.12 & 8.98 & 24.31 & 22.87 & 61.63 & \textbf{30.17} & 16.41 & 13.14 & 10.33  \\ 
 \rowcolor{lightgray!30} STADE & 2:4 & \textbf{11.38} & \textbf{10.82} & \textbf{8.42} & \textbf{22.30} & \textbf{20.52} & \textbf{46.90} & 133.17 & \textbf{15.20} & \textbf{12.52} & \textbf{10.13}  \\ 
  \midrule 
Magnitude & 4:8 & 16.83 & 15.91 & 7.32 & 181.47 & 212.46 & 614.71 & 150.43 & 115.48 & 21.18 & 36.49  \\ 
 Wanda & 4:8 & \textbf{8.57} & 8.60 & 7.00 & 14.61 & 13.78 & 32.62 & \textbf{22.25} & 13.24 & 11.12 & 9.46  \\ 
 \rowcolor{lightgray!30} STADE & 4:8 & 8.63 & \textbf{8.29} & \textbf{6.86} & \textbf{13.69} & \textbf{12.93} & \textbf{27.76} & 51.13 & \textbf{12.62} & \textbf{10.69} & \textbf{9.29}  \\ 
  \midrule 
Magnitude & 50\% & 17.29 & 16.03 & 6.83 & 205.45 & 134.28 & 174.10 & 111.22 & 54.56 & 15.22 & 49.09  \\ 
 Wanda & 50\% & \textbf{7.26} & \textbf{6.92} & 5.97 & 9.83 & 9.65 & 20.63 & \textbf{16.39} & 11.35 & 10.00 & \textbf{8.63}  \\ 
 \rowcolor{lightgray!30} STADE & 50\% & 7.43 & 6.97 & \textbf{5.95} & \textbf{9.63} & \textbf{9.47} & \textbf{18.67} & 16.90 & \textbf{11.19} & \textbf{9.60} & 8.65  \\ 
\bottomrule 
 \end{tabular}
 \caption{Perplexity on Wikitext2 for different Llama and Qwen models. C4 dataset is used as the calibration dataset during the pruning process. 2:4 and 4:8 sparsity refers to a structure pruning approach where 2/4 weights are pruned out of every 4/8 weights \cite{nvidia_prune_n_m}}
 \label{tab:main_experiment}
 \end{table*}

\section{Experiments}

\textbf{Models and Evaluation.} Most experiments are conducted using the Llama \cite{llama1,llama2,llama3} and Qwen \cite{qwen, qwen2, qwen2.5, qwen3} models. In addition, the OPT family \cite{zhang2023opt} is also evaluated due to its architectural differences such as the usage of a bias term in the linear layers, the usage of \textit{Layernorm} \cite{layernorm} and the incorporation of positional embeddings instead of rotary position embeddings \cite{roformer}.

Following previous research \cite{wanda}, C4 dataset \cite{c4dataset} is used for calibration, while raw-WikiText2 dataset \cite{wikitex2} is employed to evaluate model perplexity. Moreover, the zero-shot capabilities of the pruning methods are assessed using nine tasks from the EleutherAI LM Harness Benchmark \cite{eval-harness}. These tasks include: \textit{Boolq} \cite{clark2019boolq}, a yes/no question answering dataset containing 15,942 examples; the \textit{Recognizing Textual Entailment (RTE)} suite, which combines RTE-1 \cite{rte1}, RTE-2 \cite{rte2}, RTE-3 \cite{rte3}, and RTE-5 \cite{rte5} challenges constructed from news and Wikipedia text; \textit{HellaSwag} \cite{zellers2019hellaswag}, a challenging dataset for evaluating commonsense,; \textit{WinoGrande} \cite{ai2:winogrande}, a binary fill-in-the-blank task that requires commonsense reasoning; \textit{Arc-Easy} and \textit{Arc-Challenge} \cite{allenai:arc}, which consist of multiple-choice science questions targeting grade-school level content and are split into easy and challenging subsets, \textit{OpenBookQA} \cite{OpenBookQA2018}, a dataset that involves questions requiring multi-step reasoning, additional commonsense knowledge, and comprehensive text comprehension;
%\textit{GSM8K (Grade School Math 8K)} \cite{gsm8k}, a dataset of 8.5K high quality linguistically diverse grade school math word problems;
and \textit{MMLU} \cite{mmlu}, a multitask test consisting of multiple-choice questions from various branches of knowledge.

\textbf{Baselines.} The main experiments employ pruning methods that do not involve weight updates to corroborate our theoretical analysis. These methods include \textit{Magnitude} pruning \cite{magnitudeprune} and \textit{Wanda} \cite{wanda}. Furthermore, we also do an ablation on methods with weight updates (\textit{SparseGPT} \cite{sparsegpt}) for further insights.

\textbf{Pruning.} The pruning strategy follows a layer-wise approach, which can be easily augmented with more complex procedures that assign different weights to each layer \cite{xubesa, softprune}. The main focus is on unstructured pruning, where any weight in a matrix may be pruned. Additionally, the structured N:M pruning scenario will also be evaluated. In N:M structure pruning, out of every M weights N must be pruned \cite{prune_n_m}. In particular, the 2:4 and 4:8 structured pruning schemes proposed by Nvidia \cite{nvidia_prune_n_m} for faster inference are adopted.

\subsection{Large Language Modeling pruning}\label{sec:large_language_modeling}

Table \ref{tab:main_experiment} reports the perplexity of Llama and Qwen models with various pruning methods. Notice that \textit{STADE} outperforms the other methods consistently across the different pruning scenarios. 
These results follow our formal analysis, validating our theoretical understanding.
Notice that the LLMs in Table \ref{tab:main_experiment} use RMSNorm \cite{rmsnorm} and therefore we do not use \textit{STADE-W}.
Since no layer receives a normalized input, it is no different from standard \textit{STADE}.

 \begin{table*}[h] 
 \centering 
 \begin{tabular}{ccccccccc} 
 \toprule 
 Method & Sparsity & opt-125m & opt-350m & opt-1.3b & opt-2.7b & opt-6.7b & opt-13b & opt-30b \\ 
 \midrule 
 Dense & 0\% & 27.65 & 22.00 & 14.62 & 12.47 & 10.86 & 10.13 & 9.56 \\ 
 \midrule 
 Magnitude & 2:4 & 341.46 & 417.01 & 427.09 & 1152.92 & 264.04 & 484.64 & 1981.10  \\ 
 Wanda & 2:4 & 80.24  & 113.54  & 28.23  & 21.20  & 15.89  & \textbf{15.52 } & 13.44   \\ 
 STADE & 2:4 & 109.68  & 100.16  & \textbf{27.19} & 24.08  & 16.44  & 17.61  & 15.35  \\ 
 \rowcolor{lightgray!30} STADE-W & 2:4 & \textbf{76.08 } & \textbf{99.82} & 27.55 & \textbf{20.68} & \textbf{15.64} & 15.57 & \textbf{12.40} \\ 
  \midrule 
Magnitude & 4:8 & 169.09 & 160.73 & 240.13 & 166.93 & 196.15 & 450.06 & 564.03 \\ 
Wanda & 4:8 & 53.18 & 58.49 & 22.15 & 16.77 & 13.56 & 13.37 & 10.88  \\ 
 STADE & 4:8 & 68.19 & 57.62 & \textbf{21.34} & 17.38 & 13.79 & 14.98 & 11.42  \\ 
 \rowcolor{lightgray!30} STADE-W & 4:8 & \textbf{52.64} & \textbf{56.69} & 21.93 & \textbf{16.66} & \textbf{13.41} & \textbf{13.34} & \textbf{10.85}  \\ 
  \midrule 
Magnitude & 50\% & 193.35 & 97.78 & 1713.49 & 265.17 & 968.77 & 11609.08 & 168.09  \\ 
 Wanda & 50\% & \textbf{38.94} & 36.21 & 18.42 & 14.22 & 11.98 & \textbf{11.92} & \textbf{10.03} \\ 
 STADE & 50\% & 49.04 & 37.51 & \textbf{17.75} & 14.36 & \textbf{11.87} & 13.10 & 10.19  \\ 
 \rowcolor{lightgray!30} STADE-W & 50\% & 39.22 & \textbf{36.15} & 18.38 & \textbf{14.20} & 11.97 & 11.96 & 10.05 \\ 
\bottomrule 
 \end{tabular} 
 \caption{Perplexity on Wikitext2 with C4 as the calibration dataset.} 
 \label{tab:wikitext_opt} 
 \end{table*}

\begin{figure}[h]
\centering
\captionsetup{justification=centering}
	\begin{subfigure}{0.23\textwidth}
    	\includegraphics[width=\textwidth]{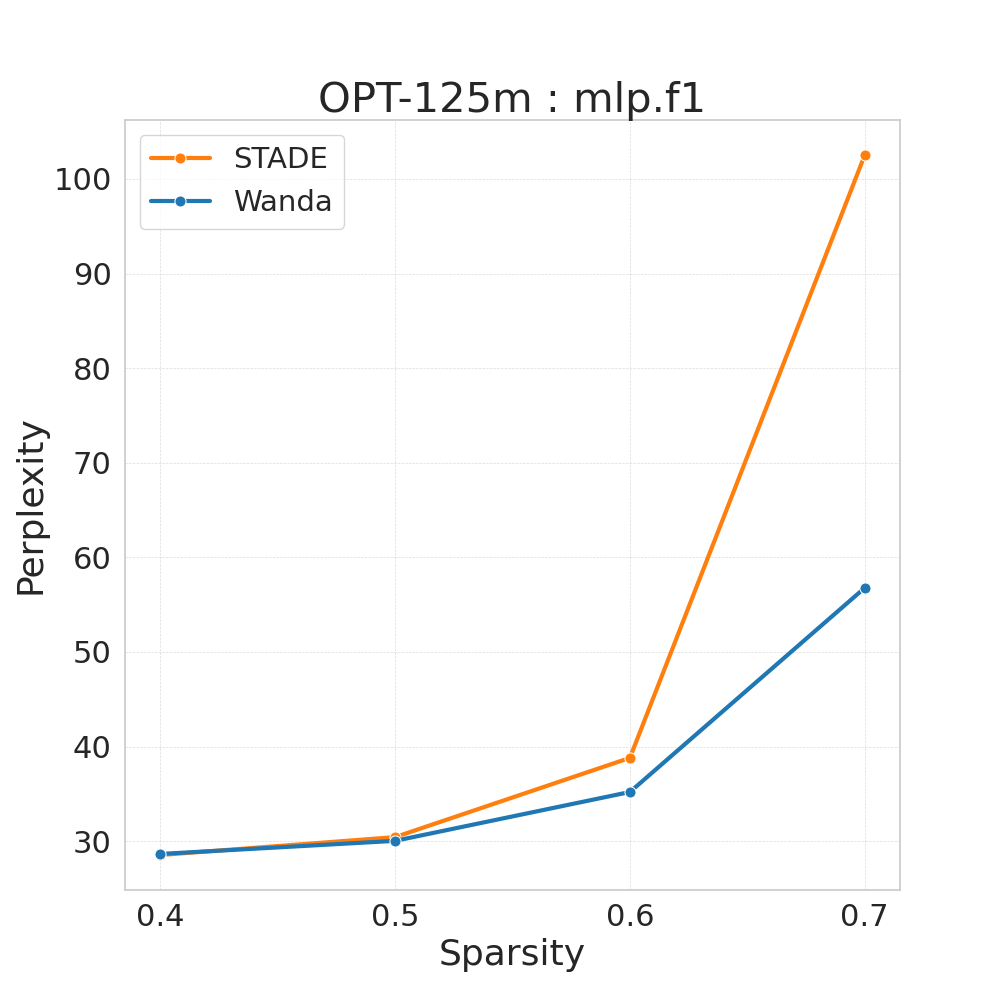} %{images/result_mlp.fc1_plots_seaborn.png}
    	\caption{Centered input \\ (mlp.f1 layer)}
    	\label{fig:first}
	\end{subfigure}
	\hfill %hspace{2cm}
	\begin{subfigure}{0.23\textwidth}
    	\includegraphics[width=\textwidth]{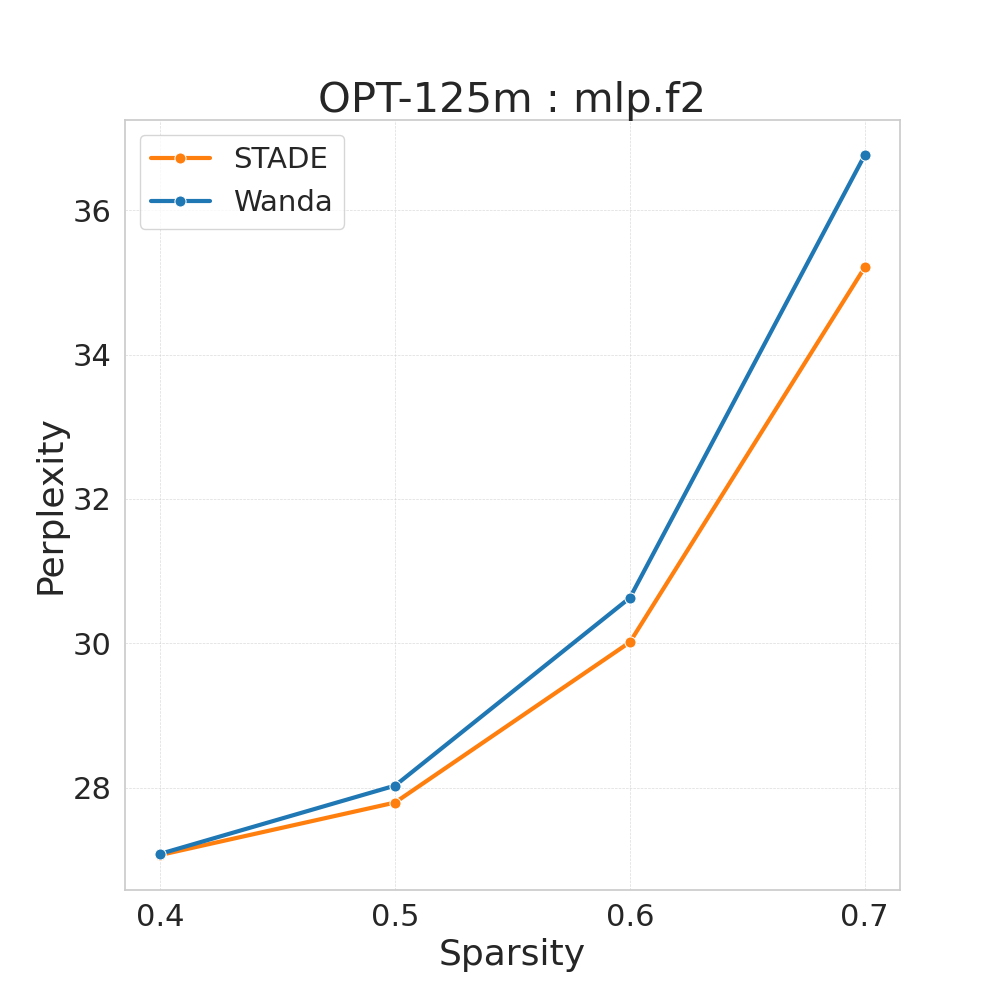} %{images/result_mlp.fc2_plots_seaborn.png}
    	\caption{Not centered input \\ (mlp.f2 layer)}
    	\label{fig:second}
	\end{subfigure}
  \caption{Perplexity comparison on OPT-125m when pruning only the specified layer type.}
  \label{fig:mlp_comparison}
\end{figure}

\subsection{Pruning requirements of different layers}\label{sec:experiments_stade_w}

In order to see the effect of normalization layer, we investigate OPT models which use \textit{Layernorm} \cite{layernorm} instead of \textit{RMSNorm} \cite{rmsnorm}. 
% Prior research \cite{xubesa} has demonstrated that different layers influence overall model performance to varying degrees when pruned. However, to the best of our knowledge, no previous work has systematically investigated whether distinct pruning metrics may be beneficial when applied to different layers (Fig. \ref{fig:mlp_comparison}).
Since the linear layers that receive the input after a \textit{Layernorm} would be centered, a small ablation is done on the difference when pruning a layer with a centered input vs an uncentered input in Figure \ref{fig:mlp_comparison}. 

The experiment show that different layers benefit from different pruning methods. In particular, when a layer receives a centered input (first layer of the MLP block), \textit{Wanda} performs better since it already assumes this scenario while \textit{STADE} approximates the mean with the inputs. However, whenever the input is not centered \textit{Wanda} is not able to keep up with \textit{STADE} (second layer of the MLP block).
This result is in line with our theoretical analysis and validates our characterization of the pruning problem.
With these finding we propose \textit{STADE-W}, a method that combines both \textit{STADE} and \textit{Wanda}. It uses \textit{Wanda} when the input is centered and \textit{STADE} otherwise. The results in Table \ref{tab:wikitext_opt} show that \textit{STADE-W} improves model performance over \textit{STADE} or \textit{Wanda} when used individually on the OPT family.

 \subsection{Zero-shot comparison}\label{sec:zero_shot_experiments}
While model perplexity serves as an important evaluation of pruning strategies, measuring prediction accuracy is equally crucial for large language models and their pruned variants. To test the impact of the different pruning methods on model accuracy, we evaluate on multiple zero-shot tasks across different datasets. The results are reported in Table \ref{tab:zero_shot}.

The results on the zero-shot task align with those observed when evaluating perplexity (Table \ref{tab:main_experiment}).
\textit{STADE} method demonstrates competitive performance across a range of models and tasks.

\begin{table}[H]
 \centering
 \begin{tabular}{ccccc}
 \toprule
 \multirow{2}*{Method} & Weight & \multirow{2}*{Sparsity} & Qwen3 & Qwen3 \\
  & Update &  & 1.7B & 8B \\
 \midrule
 Dense & \xmark & 0\% & 16.67 & 9.72 \\
  \midrule
Magnitude & \xmark & 2:4 & 1808.24 & 294.48  \\
 Wanda & \xmark & 2:4 & 61.63 & 16.41  \\
 SparseGPT & \cmark & 2:4 & \textbf{31.74} & \textbf{14.48}  \\
 SparseGPT & \multirow{2}*{\xmark} & \multirow{2}*{2:4} & \multirow{2}*{51.75} & \multirow{2}*{\underline{14.99}}  \\
 (w/o update) & & &  &  \\
 \rowcolor{lightgray!30} STADE & \xmark & 2:4 & \underline{46.90} & 15.20  \\
  \midrule
Magnitude & \xmark & 4:8 & 614.71 & 115.48  \\
 Wanda & \xmark & 4:8 & 32.62 & 13.24  \\
 SparseGPT & \cmark & 4:8 & \textbf{25.38} & \underline{12.65}  \\
 SparseGPT & \multirow{2}*{\xmark} & \multirow{2}*{4:8} & \multirow{2}*{28.80} & \multirow{2}*{13.02}  \\
 (w/o update) & & &  &  \\
 \rowcolor{lightgray!30} STADE & \xmark & 4:8 & \underline{27.76} & \textbf{12.62}   \\
 \midrule
Magnitude & \xmark & 50\% & 174.10 & 54.56  \\
 Wanda & \xmark & 50\% & \underline{20.63} & \underline{11.35}  \\
 SparseGPT & \cmark & 50\% & 23.73 & 11.49  \\
 SparseGPT & \multirow{2}*{\xmark} & \multirow{2}*{50\%} & \multirow{2}*{22.89} & \multirow{2}*{11.80}  \\
 (w/o update) & & &  &  \\
 \rowcolor{lightgray!30} STADE & \xmark & 50\% & \textbf{18.67} & \textbf{11.19}  \\
\bottomrule
 \end{tabular}
 \caption{Perplexity comparison with pruning methods that update weights (\textit{SparseGPT}).}
 \label{tab:sparsegpt_no_update}
 \end{table} \begin{table*}[h] 
 \centering 
 \begin{tabular}{ccccccc} 
 \toprule 
 Method & Sparsity & Qwen3-0.6B & Qwen3-1.7B & Qwen3-4B & Qwen3-8B & Qwen3-14B \\ 
 \midrule 
 Dense & 0\% & 45.73\% & 56.42\% & 63.52\% & 66.63\% & 69.48\% \\ 
 \midrule 
 Magnitude & 50\% & 32.91\% & 33.63\% & 35.40\% & 41.11\% & 60.65\%  \\ 
 Wanda & 50\% & 40.25\% & 49.85\% & \textbf{57.66\%} & 62.48\% & 67.23\%  \\ 
 \rowcolor{lightgray!30} STADE & 50\% & \textbf{40.36\%} & \textbf{50.38\%} & 57.10\% & \textbf{62.99\%} & \textbf{67.64\%}  \\ 
  \midrule 
Magnitude & 2:4 & 30.21\% & 33.09\% & 33.11\% & 33.62\% & 48.54\%  \\ 
 Wanda & 2:4 & 33.28\% & 38.57\% & \textbf{47.28\%} & 54.71\% & 59.81\%  \\ 
 \rowcolor{lightgray!30} STADE & 2:4 & \textbf{33.85\%} & \textbf{39.72\%} & 43.79\% & \textbf{56.62\%} & \textbf{59.94\%}  \\ 
  \midrule 
Magnitude & 4:8 & 31.17\% & 33.75\% & 35.43\% & 34.26\% & 54.53\%  \\ 
 Wanda & 4:8 & \textbf{35.49\%} & 43.21\% & \textbf{53.85\%} & 60.17\% & 63.33\%  \\ 
 \rowcolor{lightgray!30} STADE & 4:8 & 34.89\% & \textbf{43.26\%} & 47.99\% & \textbf{60.52\%} & \textbf{63.69\%}  \\ 
\bottomrule 
 \end{tabular} 
 \caption{Zero shot accuracy averaged over 9 individual tasks (\textit{Arc-Challenge}, \textit{Arc-Easy}, \textit{Boolq}, \textit{HellaSwag}, \textit{OpenBookQA}, \textit{RTE-3}, \textit{WinoGrande} and \textit{MMLU}. The results for each individual tasks can be found in the Appendix.} 
 \label{tab:zero_shot} 
 \end{table*}

\subsection{Weight update importance}

\textit{SparseGPT} is a pruning criterion that despite being comparable to \textit{Wanda} and in some cases even outperforming it, it is known to be slower and more computationally demanding than other baselines. In this section, we compare the performance of \textit{SparseGPT} against \textit{STADE}, with a particular focus on the critical importance of its weight update mechanism.

Table \ref{tab:sparsegpt_no_update} shows that even though \textit{SparseGPT} is competitive, it is not always the best performing method.
In particular, it is a competitive for scenarios where the weight selection is more restricted, i.e., it gets better with smaller models and with more constrained scenarios such as 2:4 pruning.
Nevertheless, when pruning bigger models in more unstructured scenarios, \textit{STADE} is able to show SOTA performance.
Moreover, the efficacy of \textit{SparseGPT} comes from the weight update and not from the weight selection criterion as shown with the \textit{SparseGPT (w/o update)} ablation.

Notice that our theoretical analysis of the pruning problem presented focuses exclusively on cases without weight updates. The experimental findings corroborate that the performance of pruning methods can be improved by applying weight updates to the remaining parameters after pruning. These updates help to counterbalance the negative effects associated with the removal of weights, thereby better preserving the model's predictive capability. This suggests that beyond the initial selection of weights to prune, the adjustment of the unpruned weights is a crucial factor in achieving optimal performance.

\subsection{Bias usage ablation}

\textit{STADE} method updates the bias term when pruning the models. In models like OPT which already have bias, this is reasonable. However, Llama and Qwen models do not have originally a bias term and therefore, this would results in adding a new bias term which could be considered as adding some extra weight variables. This could be considered an unfair advantage when compared to the other methods. To investigate this, a small ablation is done where we do not update the bias term.

\begin{table}[h]
    \centering
    \begin{tabular}{cccc}
        \toprule
        \multirow{2}*{Model} & \multirow{2}*{Sparsity} & \multirow{2}*{STADE} & STADE \\
         & & & (w/o bias) \\
         \midrule
        Llama-7B & 0.5 & \textbf{11.38} & \textbf{11.38} \\
        Llama2-7B & 0.5 & \textbf{10.82} & \textbf{10.82} \\
        Llama2-13B & 0.5 & \textbf{8.42} & \textbf{8.42} \\
        Llama3-8B & 0.5 & \textbf{22.30} & 22.37\\
        Llama3.1-8B & 0.5 & \textbf{20.52} & \textbf{20.52} \\
        Qwen3-1.7B & 0.5 & \textbf{46.90} & \textbf{46.90} \\
        Qwen3-4B & 0.5 & 133.17 & \textbf{131.50} \\
        Qwen3-8B & 0.5 & \textbf{15.20} & \textbf{15.20} \\
        Qwen3-14B & 0.5 & 12.52 & \textbf{12.51}\\
        Qwen3-32B & 0.5 & \textbf{10.13} & \textbf{10.13} \\
        \bottomrule
    \end{tabular}
    \caption{Ablation on the importance of the bias update in \textit{STADE} algorithm.}
    \label{tab:importance_of_bias}
\end{table}

The results shown in Tab. \ref{tab:importance_of_bias} demonstrate that there is little to no difference when adding the bias term and if one wants to remove this term the results are almost identical. Empirically we observe that for any layer, the sum of the absolute terms of the bias is always smaller than $10^{-2}$, which explains why removing it has little to no impact.

\section{Future Work}

The exploration of various pruning criterions has revealed that no single method is universally optimal for every layer within a deep neural network. Future research should aim to deepen the understanding of how different layers and network depths interact with distinct pruning criteria, potentially leading to adaptive, layer-specific pruning strategies. In addition, investigating the benefits of pruning each layer with different sparsity ratios could further enhance model efficiency and performance, representing another promising direction for future work. Furthermore, methods such as \textit{SparseGPT} demonstrate that incorporating weight updates for unpruned parameters can significantly enhance performance, suggesting that further investigation into efficient weight update mechanisms may yield substantial benefits.

In our work we focus on medium to small model sizes, as we are interested in their capabilities on many resource-constrained scenarios.
Nevertheless, evaluating these pruning methods on larger models and other architectures (\textit{mixture of experts} \cite{mixtral}) would help to assess the scalability and effectiveness of the proposed techniques in on large-scale settings.
%Finally, a systematic study that combines and optimizes different pruning metrics for specific layers or blocks may pave the way for more robust and efficient model compression techniques, thereby facilitating the deployment of large language models on resource-constrained devices.

\section{Conclusion}

This work presents a comprehensive analysis of optimal weight pruning in neural networks and provides a theoretical framework that explains why \textit{Wanda} is effective in many common deep learning scenarios. It demonstrates that while \textit{Wanda} performs optimally in layers with centered inputs, its effectiveness diminishes in layers that receive uncentered inputs. In response to these observations, we propose a new pruning criterion (\textit{STADE}) that handles this scenario. We demonstrate theoretically and empirically that \textit{STADE} outperforms \textit{Wanda} for uncentered inputs.

We also observe that different layers have different input statistics and therefore the optimal pruning criterion might change between layers. Building upon these insights, we introduce \textit{STADE-W}, which dynamically combines \textit{Wanda} and \textit{STADE} based on the input statistics of each layer, making it, to the best of current knowledge, the first pruning method that employs different pruning criterions for different layers. We do extensive experiments on Qwen, Llama and Open Pre-trained Transformers models. We not only evaluate perplexity but also zero-shot performance. The results validate our theoretical analysis and reveal that pruning effectiveness varies according to the input characteristics of each layer. Notably, our method achieves state-of-the-art performance for the pruning problem. Moreover, our experiments demonstrate that incorporating weight update mechanisms (as exemplified by \textit{SparseGPT}) can improve performance, further highlighting the benefits of updating the unpruned weights and a future research direction.

All together, these contributions not only advance the understanding of pruning strategies but also offer a new robust method for reducing the computational demands of large language models without significant performance loss. The insights provided herein pave the way for more efficient deployment of large-scale models in resource-constrained environments.

% This work performs a thorough analysis on how to optimally prune weights from a neural network, from which it is proved that \textit{Wanda} is the optimal in a lot of common scenarios in Deep Learning.
% However, it also noticed that \textit{Wanda} is suboptimal in some of the layers from a transformer, in particular, for those layers that receive a not centered/biased input (output layer of multi-head attention and second layer of the MLP). After analyzing those cases, an optimal pruning metric (\textit{STD}) based on standard deviation of the inputs is deduced, which handles optimally the cases where the input is biased.
% With this new knowledge,a new pruning method \textit{STADE}is defined, which combines both pruning metrics depending on the input statistics.
% As far as we know, this is the first pruning method to change between pruning metrics.
% To validate the theoretical analysis, \textit{STD} and \textit{Wanda} metrics are compared showing that different layers perform better with different pruning metrics.
% \textit{STADE} is then compared with other state-of-the-art pruning methods showing comparable performance for the Llama and OPT models, and outperforming them for the newer and more capable versions of them (Llama-3).

% \section{Acknowledgments}

% This work was supported by the \textit{Information Science and Machine Learning Lab} (\textbf{ISMLL}) in University of Hildesheim and \textit{VolksWagen Financial Services Data Analytics Research Center} (\textbf{VWFS DARC}).

\bibliography{aaai2026}

\section{Appendix}

\subsection{Pruning time}

Different pruning methods take different time to calculate their corresponding pruning scores.
We report in Table \ref{tab:pruning_time} the pruning time for the different methods in different pruning scenarios.

\begin{table}[h]
    \centering
    \begin{tabular}{c|cccc}
        \toprule
        Sparsity& Wanda& STADE & STADE-W &SparseGPT\\
        \midrule
        50\%& 72,89& 72,02& 74,55&222,31\\
        2:4& 77,87 &74,80& 73,04&204,52\\
        4:8& 70,10& 73,39 & 71,51 &215,36\\
        \bottomrule
    \end{tabular}
    \caption{Pruning time comparison in seconds for different methods on Llama-3.2-1B.}
    \label{tab:pruning_time}
\end{table}

\subsection{Implementation details}

When estimating the mean and the standard deviation, loading the full data results in an Out-Of-Memory error with our computational resources.
Therefore, the mean and standard deviation is calculated using a in a moving fashion.\\

Calculating the sum and square of sums to later approximate the final results, leads to too high values at which point the new instances do not add any values to the value stored.
In order to avoid that, the mean and standard deviation is calculated in each iteration as follows:

\begin{algorithm}[h]
\caption{Mean and variance calculation}
\label{alg:mean_and_std_calculation}
\textbf{Input}: dataloader \textit{D}\\
\textbf{Output}: mean $\mu$, var $\mathbb{V}$
\begin{algorithmic}[1] %[1] enables line numbers
\STATE Let $N=0$, $\mu=0$ and $\mathbb{V}=0$.
\FOR{$x_{batch}$ in \textit{D}}
\STATE $N_{new}=N+len(x_{batch})$
\STATE $\mu_{new}=\frac{\mu*N}{N_{new}} + \frac{sum(x\_batch)}{N_{new}}$
\STATE $\mathbb{V}=\frac{(N-1)*\mathbb{V}}{N_{new}-1} + \frac{N*\mu^2}{(N_{new}-1)} - \frac{N_{new}*\mu_{new}^2}{(N_{new}-1)} + \frac{sum(x_{batch}^2)}{N_{new}-1}$
\STATE $N=N_{new}$
\STATE $\mu=\mu_{new}$
\ENDFOR
\STATE \textbf{return} $\mu$, $\mathbb{V}$
\end{algorithmic}
\end{algorithm}

\subsection{Training details}

While the pruning methods shown don't have hyperparameter to tuned, there are some training details that we would like to mention:

\begin{itemize}
    \item \textbf{C4 calibration dataset}: Following \textit{Wanda}, when using C4 dataset only the file \textit{'en/c4-train.00000-of-01024.json.gz'} is used during pruning to speed up the process. the full dataset can be fined in \textit{'https://huggingface.co/datasets/allenai/c4/tree/main/en'}.
    \item \textbf{Sequence length}: Some LLMs allow over 10k context window. In order to run the models under our personal hardware constrains, we reduce the sequence length to 2048 in order to test more models. This is both applied during pruning and evaluation.
\end{itemize}

\subsection{Intuitive explanation of STADE}

In this section we formulate an easy and intuitive explanation for \textit{STADE}.
We will assume that the input multivariate distribution $X \in \mathbb{R}^{2}$ is normally distributed, i.e., $X_i \sim \mathcal{N}(\mu_i,\sigma_i)$. 
\textbf{Notice that STADE does not require the input to be normally distributed,} this is just a simplification for the sake of the explanation.
In the same way as in the Methodology section, we will consider the pruning problem for one column. In this case the corresponding output of the linear layer ($\hat{y}$) can be calculated as:

\begin{equation}\label{eq:problem_definition}
    \begin{split}
        \hat{y} &= B + x_1 W_1 + x_2 W_2 =\\
        &=B + (\mu_1 + \epsilon_1 \sigma_1) W_1 + (\mu_2 + \epsilon_2 \sigma_2) W_2\\
        &= (B + \mu_1 + \mu_2) + \sigma_1 W_1 \epsilon_1 + \sigma_2 W_2 \epsilon_2
    \end{split}
\end{equation}

Notice that $\epsilon_1, \epsilon_2 \sim \mathcal{N}(0, 1)$ and therefore will affect the same way when pruning the weight.
However, when deciding whether to prune $W_1$ or $W_2$, it's not only about the value of the weight but also about the standard deviation of the corresponding input.
This is due to the fact that the mean of the input can be added to the bias and therefore omitted when pruning the weights.

\subsection{STADE* variation derivation}

Not all models use a bias term in their linear layers. Therefore, we consider a variation of \textit{STADE} without the possibility of updating the bias (\textit{STADE*}), i.e., the linear layer to prune has no bias term and the pruning method is not allowed to add a bias term in order to keep the model structure. To do so we expand on the previous derivations from the main paper as follows:

\begin{equation}\label{eq:bias_wanda_solution}
	\begin{split}
    	&\min_{W^*,0} \mathbb{E}[\left( (B + \sum_{i=1}^M X_i W_i) - (B^* + \sum_{i=1}^M X_i W^*_i)\right)^2] \\
     	&= \min_{j,0} (B - B^*)^2 + 2(B - B^*) \mu_j W_j + (\sigma_j^2 + \mu_j^2)W_j^2\\
     	&= \min_{j} (\sigma_j^2 + \mu_j^2)W_j^2\\
        &\approx \left[||X_{:,j} - \frac{1}{N}\sum_{n=1}^N X_{n,j}||_2^2 + (\frac{1}{N}\sum_{n=1}^N X_{n,j})^2\right] |W_{i,j}|^2
	\end{split}
\end{equation}

 \begin{table*}[h] 
 \centering 
 \begin{tabular}{cccccccc} 
 \toprule 
 Method & Sparsity & Qwen3-0.6B & Qwen3-1.7B & Qwen3-4B & Qwen3-8B & Qwen3-14B & Qwen3-32B \\ 
 \midrule 
 Dense & 0\% & 20.95 & 16.67 & 13.64 & 9.72 & 8.64 & 7.60\\
 \midrule 
 Magnitude & 2:4 & 85481.66 & 1808.24 & 1970.45 & 294.48 & 38.58 & 29.89  \\ 
 Wanda & 2:4 & \underline{190.03} & 61.63 & \underline{30.17} & 16.41 & 13.14 & \underline{10.33}  \\ 
 \rowcolor{lightgray!30} STADE & 2:4 & 13785.28 & \underline{46.90} & 133.17 & \underline{15.20} & 12.52 & \textbf{10.13}  \\ 
 \rowcolor{lightgray!30} STADE (w/o bias) & 2:4 & 13785.28 & \underline{46.90} & 131.50 & \underline{15.20} & 12.51 & \textbf{10.13}  \\ 
 STADE* & 2:4 & 193.29 & 60.66 & 30.25 & 16.41 & 13.10 & -  \\ 
 STADE-W & 2:4 & 171.60 & 47.24 & 32.37 & 15.54 & 12.57 & -  \\ 
 SparseGPT & 2:4 & \textbf{91.05} & \textbf{31.74} & \textbf{21.32} & \textbf{14.48} & \underline{12.47} & -  \\ 
 SparseGPT (no update) & 2:4 & 6278.69 & 51.75 & 86.57 & 14.99 & \textbf{12.18} & -  \\ 
  \midrule 
Magnitude & 4:8 & 99815.32 & 614.71 & 150.43 & 115.48 & 21.18 & 36.49  \\ 
 Wanda & 4:8 & \underline{73.71} & 32.62 & \underline{22.25} & 13.24 & 11.12 & \underline{9.46}  \\ 
 \rowcolor{lightgray!30} STADE & 4:8 & 304.30 & \underline{27.76} & 51.13 & \textbf{12.62} & 10.69 & \textbf{9.29}  \\ 
 \rowcolor{lightgray!30} STADE (w/o bias) & 4:8 & 304.30 & 27.79 & 52.95 & \underline{12.63} & \underline{10.68} & \textbf{9.29}  \\ 
 STADE* & 4:8 & 74.73 & 32.46 & 22.50 & 13.24 & 11.12 & -  \\ 
 STADE-W & 4:8 & 77.89 & 28.67 & 22.57 & 12.71 & \textbf{10.67} & -  \\ 
 SparseGPT & 4:8 & \textbf{60.55} & \textbf{25.38} & \textbf{18.64} & 12.65 & 11.16 & -  \\ 
 SparseGPT (no update) & 4:8 & 513.65 & 28.80 & 51.60 & 13.02 & 10.71 & -  \\ 
  \midrule 
Magnitude & 50\% & 1455.57 & 174.10 & 111.22 & 54.56 & 15.22 & 49.09  \\ 
 Wanda & 50\% & 34.20 & 20.63 & 16.39 & 11.35 & 10.00 & \textbf{8.63}  \\ 
 \rowcolor{lightgray!30} STADE & 50\% & \underline{34.01} & \underline{18.67} & 16.90 & \underline{11.19} & \underline{9.60} & \underline{8.65}  \\ 
 \rowcolor{lightgray!30} STADE (w/o bias) & 50\% & 34.04 & \textbf{18.66} & 16.90 & \underline{11.19} & \underline{9.60} & \underline{8.65}  \\ 
 STADE* & 50\% & 34.06 & 20.57 & \underline{16.39} & 11.35 & 10.01 & -  \\ 
 STADE-W & 50\% & \textbf{33.96} & 18.98 & \textbf{16.04} & \textbf{11.10} & \textbf{9.54} & -  \\ 
 SparseGPT & 50\% & 34.14 & 23.73 & 17.39 & 11.49 & 10.08 & -  \\ 
 SparseGPT (no update) & 50\% & 89.12 & 22.89 & 20.03 & 11.80 & 9.95 & -  \\
\bottomrule 
 \end{tabular} 
 \caption{Wikitext perplexity for the Qwen family. Notice that \textit{STADE-W} here is applied after the \textit{RMSnorm} layers. As explained in the Methodology, Qwen3 uses \textit{RMSnorm} which does not normalize the inputs and therefore it is not applicable as it was with the OPT family. We observe huge spikes for Qwen3-0.6B and Qwen3-4B. To the best of our knowledge the only difference with the other models is the usage of tie-embeddings. Nevertheless, Qwen3-1.7B also uses them and doesn't exhibit those spikes. We were not able to identify the source behind it. We will investigate it in our future work.} 
 \label{tab:wikitext_qwen} 
 \end{table*}

We originally wanted to include this in the main paper. Nevertheless, standard \textit{STADE} had better performance than \textit{STADE*} even when not updating the bias, i.e., the pruning criterion even though locally optimal (optimal for the linear layer pruning) is not optimal globally (for the model pruning). This can be observed in the following tables. The only cases, where it outperforms \textit{STADE} is when \textit{STADE} has a big spike/jump on the perplexity. It seems that it is performing worse in general, but it is always to have consistent results avoiding the huge spikes. We will investigate this phenomena more in the future work.

\subsection{Additional experiments}\label{sec:appendix_zero_shot}

Tables \ref{tab:wikitext_qwen} to \ref{tab:mmlu} show experiments on more models and additional pruning metrics measuring both perplexity and zero-shot performance.

 \begin{table*}[h] 
 \centering 
 \begin{tabular}{ccccccc} 
 \toprule 
 Method & Sparsity & Qwen3-0.6B & Qwen3-1.7B & Qwen3-4B & Qwen3-8B & Qwen3-14B \\ 
 \midrule 
 Dense & 0\% & 31.40\% & 39.76\% & 50.77\% & 55.80\% & 58.62\% \\ 
 \midrule 
 Magnitude & 50\% & 20.90\% & 19.11\% & 22.70\% & 28.07\% & 48.81\%  \\ 
 Wanda & 50\% & 23.38\% & 30.46\% & 39.76\% & 50.85\% & \textbf{55.20\%}  \\ 
 \rowcolor{lightgray!30} STD (w/o bias) & 50\% & \textbf{23.98\%} & \textbf{33.53\%} & \textbf{41.72\%} & \textbf{51.88\%} & 55.03\%  \\ 
  \midrule 
Magnitude & 2:4 & 20.65\% & 20.56\% & 22.44\% & 18.69\% & 37.12\%  \\ 
 Wanda & 2:4 & 19.71\% & 19.71\% & \textbf{32.17\%} & 38.65\% & 42.24\%  \\ 
 \rowcolor{lightgray!30} STD (w/o bias) & 2:4 & \textbf{21.67\%} & \textbf{20.90\%} & 29.86\% & \textbf{42.32\%} & \textbf{44.54\%}  \\ 
  \midrule 
Magnitude & 4:8 & 20.56\% & 21.50\% & 23.46\% & 19.45\% & 44.37\%  \\ 
 Wanda & 4:8 & 19.71\% & 24.91\% & \textbf{38.05\%} & 46.16\% & 50.60\%  \\ 
 \rowcolor{lightgray!30} STD (w/o bias) & 4:8 & \textbf{21.25\%} & \textbf{26.37\%} & 33.11\% & \textbf{47.95\%} & \textbf{51.96\%}  \\ 
\bottomrule 
 \end{tabular} 
 \caption{Zero-shot performance on Arc Challenge \cite{allenai:arc}.} 
 \label{tab:arc challenge} 
 \end{table*}

\begin{table*}[h] 
 \centering 
 \begin{tabular}{ccccccc} 
 \toprule 
 Method & Sparsity & Qwen3-0.6B & Qwen3-1.7B & Qwen3-4B & Qwen3-8B & Qwen3-14B \\ 
 \midrule 
 Dense & 0\% & 60.90\% & 72.22\% & 80.51\% & 83.46\% & 84.22\% \\ 
 \midrule 
 Magnitude & 50\% & 28.75\% & 34.01\% & 47.94\% & 58.12\% & 76.47\%  \\ 
 Wanda & 50\% & 48.65\% & 62.12\% & \textbf{72.90\%} & 80.09\% & 81.31\%  \\ 
 \rowcolor{lightgray!30} STD (w/o bias) & 50\% & \textbf{48.70\%} & \textbf{64.31\%} & 72.39\% & \textbf{80.43\%} & \textbf{81.94\%}  \\ 
  \midrule 
Magnitude & 2:4 & 26.52\% & 28.91\% & 33.46\% & 36.57\% & 63.09\%  \\ 
 Wanda & 2:4 & \textbf{32.79\%} & 47.35\% & \textbf{59.26\%} & 72.69\% & \textbf{72.47\%}  \\ 
 \rowcolor{lightgray!30} STD (w/o bias) & 2:4 & 27.86\% & \textbf{49.71\%} & 50.17\% & \textbf{74.03\%} & 71.42\%  \\ 
  \midrule 
Magnitude & 4:8 & 27.48\% & 30.51\% & 42.63\% & 40.32\% & 72.35\%  \\ 
 Wanda & 4:8 & \textbf{40.74\%} & 56.65\% & 6\textbf{7.89\%} & 77.23\% & 78.41\%  \\ 
 \rowcolor{lightgray!30} STD (w/o bias) & 4:8 & 31.65\% & \textbf{58.25\%} & 57.58\% & \textbf{77.78\%} & \textbf{78.87\%}  \\ 
\bottomrule 
 \end{tabular} 
 \caption{Zero-shot performance on Arc Easy \cite{allenai:arc}.} 
 \label{tab:arc_easy} 
 \end{table*}

\begin{table*}[h] 
 \centering 
 \begin{tabular}{ccccccc} 
 \toprule 
 Method & Sparsity & Qwen3-0.6B & Qwen3-1.7B & Qwen3-4B & Qwen3-8B & Qwen3-14B \\ 
 \midrule 
 Dense & 0\% & 64.53\% & 77.46\% & 85.11\% & 86.64\% & 89.33\% \\ 
 \midrule 
 Magnitude & 50\% & 46.36\% & 46.94\% & 38.32\% & 52.32\% & 79.60\%  \\ 
 Wanda & 50\% & \textbf{62.20\%} & \textbf{73.61\%} & \textbf{82.51\%} & \textbf{84.86\%} & 88.07\%  \\ 
 \rowcolor{lightgray!30} STD (w/o bias) & 50\% & 62.08\% & 73.43\% & 80.52\% & 84.68\% & \textbf{88.20\% } \\ 
  \midrule 
Magnitude & 2:4 & 38.65\% & 50.28\% & 46.94\% & 43.33\% & 65.78\%  \\ 
 Wanda & 2:4 & 46.76\% & 63.85\% & \textbf{73.39\%} & \textbf{82.17\%} & 85.81\%  \\ 
 \rowcolor{lightgray!30} STD (w/o bias) & 2:4 & \textbf{52.75\%} & \textbf{67.22\%} & 70.18\% & 81.68\% & \textbf{86.64\%}  \\ 
  \midrule 
Magnitude & 4:8 & 42.51\% & 51.10\% & 42.08\% & 41.28\% & 68.59\%  \\ 
 Wanda & 4:8 & 48.44\% & \textbf{70.83\%} & \textbf{78.41\%} & \textbf{85.11\%} & \textbf{87.43\%}  \\ 
 \rowcolor{lightgray!30} STD (w/o bias) & 4:8 & \textbf{57.68\%} & 66.33\% & 74.50\% & 84.92\% & \textbf{87.43\%}  \\ 
\bottomrule 
 \end{tabular} 
 \caption{Zero-shot performance on Boolq \cite{clark2019boolq}.} 
 \label{tab:boolq} 
 \end{table*}

  \begin{table*}[h] 
 \centering 
 \begin{tabular}{ccccccc} 
 \toprule 
 Method & Sparsity & Qwen3-0.6B & Qwen3-1.7B & Qwen3-4B & Qwen3-8B & Qwen3-14B \\ 
 \midrule 
 Dense & 0\% & 37.55\% & 46.12\% & 52.27\% & 57.14\% & 60.97\% \\ 
 \midrule 
 Magnitude & 50\% & 26.07\% & 28.16\% & 29.79\% & 34.51\% & 49.76\%  \\ 
 Wanda & 50\% & 32.24\% & 38.56\% & 45.03\% & 50.08\% & 55.11\%  \\ 
 \rowcolor{lightgray!30} STD (w/o bias) & 50\% & \textbf{32.86\%} & \textbf{40.26\%} & \textbf{45.92\%} & \textbf{51.65\%} & \textbf{56.66\%}  \\ 
  \midrule 
Magnitude & 2:4 & 25.70\% & 26.56\% & 27.03\% & 26.98\% & 42.16\%  \\ 
 Wanda & 2:4 & \textbf{27.17\%} & 29.90\% & \textbf{37.46\%} & 42.35\% & 48.00\%  \\ 
 \rowcolor{lightgray!30} STD (w/o bias) & 2:4 & 26.51\% & \textbf{31.08\%} & 36.22\% & \textbf{43.76\%} & \textbf{49.53\%}  \\ 
  \midrule 
Magnitude & 4:8 & 26.21\% & 26.89\% & 30.43\% & 28.04\% & 45.68\%  \\ 
 Wanda & 4:8 & \textbf{29.09\%} & 33.91\% & \textbf{41.22\%} & 46.26\% & 51.69\%  \\ 
 \rowcolor{lightgray!30} STD (w/o bias) & 4:8 & 28.27\% & \textbf{35.16\%} & 40.10\% & \textbf{47.91\%} & \textbf{53.08\%}  \\ 
\bottomrule 
 \end{tabular} 
 \caption{Zero-shot performance on HellaSwag \cite{zellers2019hellaswag}.} 
 \label{tab:hellaswag} 
 \end{table*}

  \begin{table*}[h] 
 \centering 
 \begin{tabular}{ccccccc} 
 \toprule 
 Method & Sparsity & Qwen3-0.6B & Qwen3-1.7B & Qwen3-4B & Qwen3-8B & Qwen3-14B \\ 
 \midrule 
 Dense & 0\% & 21.00\% & 28.40\% & 29.20\% & 31.00\% & 35.00\% \\ 
 \midrule 
 Magnitude & 50\% & 15.00\% & 13.40\% & 13.80\% & 18.60\% & 30.40\%  \\ 
 Wanda & 50\% & 16.60\% & 21.20\% & 26.20\% & 28.60\% & 33.00\%  \\ 
 \rowcolor{lightgray!30} STD (w/o bias) & 50\% & \textbf{18.00\%} & \textbf{22.80\%} & \textbf{26.80\%} & \textbf{30.60\%} & \textbf{33.60\%}  \\ 
  \midrule 
Magnitude & 2:4 & 14.00\% & 13.80\% & 13.00\% & 14.80\% & 26.20\%  \\ 
 Wanda & 2:4 & 13.00\% & 13.60\% & 22.20\% & 23.00\% & 28.60\%  \\ 
 \rowcolor{lightgray!30} STD (w/o bias) & 2:4 & \textbf{15.40\%} & \textbf{15.40\%} & \textbf{22.80\%} & \textbf{23.20\%} & \textbf{30.20\%}  \\ 
  \midrule 
Magnitude & 4:8 & 13.20\% & 14.40\% & 14.20\% & 15.40\% & 28.20\%  \\ 
 Wanda & 4:8 & 15.60\% & 17.40\% & \textbf{24.80\%} & 26.40\% & 31.40\%  \\ 
 \rowcolor{lightgray!30} STD (w/o bias) & 4:8 & \textbf{15.80\%} & \textbf{17.80\%} & 22.80\% & \textbf{28.40\%} & \textbf{31.60\%}  \\ 
\bottomrule 
 \end{tabular} 
 \caption{Zero-shot performance on OpenbookQA \cite{OpenBookQA2018}.} 
 \label{tab:OpenbookQA} 
 \end{table*}

  \begin{table*}[h] 
 \centering 
 \begin{tabular}{ccccccc} 
 \toprule 
 Method & Sparsity & Qwen3-0.6B & Qwen3-1.7B & Qwen3-4B & Qwen3-8B & Qwen3-14B \\ 
 \midrule 
 Dense & 0\% & 54.15\% & 70.76\% & 75.81\% & 78.34\% & 77.62\% \\ 
 \midrule 
 Magnitude & 50\% & 51.26\% & 52.71\% & 51.62\% & 52.71\% & 69.31\%  \\ 
 Wanda & 50\% & \textbf{54.15\%} & \textbf{70.04\%} & \textbf{72.92\%} & 70.04\% & 81.23\%  \\ 
 \rowcolor{lightgray!30} STD (w/o bias) & 50\% & 51.26\% & 67.15\% & 67.51\% & \textbf{70.76\%} & \textbf{82.31\%}  \\ 
  \midrule 
Magnitude & 2:4 & 42.96\% & 49.46\% & 47.29\% & 52.71\% & 48.74\%  \\ 
 Wanda & 2:4 & 51.62\% & 52.35\% & \textbf{55.96\%} & 61.73\% & \textbf{70.40\%}  \\ 
 \rowcolor{lightgray!30} STD (w/o bias) & 2:4 & \textbf{53.07\%} & \textbf{52.71\%} & 54.87\% & \textbf{69.31\%} & 65.70\%  \\ 
  \midrule 
Magnitude & 4:8 & 46.21\% & 51.26\% & 52.71\% & 52.35\% & 54.87\%  \\ 
 Wanda & 4:8 & \textbf{53.07\%} & 52.35\% & \textbf{71.12\%} & \textbf{72.92\%} & \textbf{70.04\%}  \\ 
 \rowcolor{lightgray!30} STD (w/o bias) & 4:8 & 47.65\% & \textbf{53.07\%} & 55.23\% & 70.76\% & 69.68\%  \\ 
\bottomrule 
 \end{tabular} 
 \caption{Zero-shot performance on RTE \cite{rte5}.} 
 \label{tab:rte} 
 \end{table*}

  \begin{table*}[h] 
 \centering 
 \begin{tabular}{ccccccc} 
 \toprule 
 Method & Sparsity & Qwen3-0.6B & Qwen3-1.7B & Qwen3-4B & Qwen3-8B & Qwen3-14B \\ 
 \midrule 
 Dense & 0\% & 56.20\% & 60.93\% & 66.06\% & 67.72\% & 72.85\% \\ 
 \midrule 
 Magnitude & 50\% & 49.88\% & 51.62\% & 51.62\% & 55.09\% & 64.80\%  \\ 
 Wanda & 50\% & 54.06\% & \textbf{57.38\%} & 62.12\% & \textbf{69.61\%} & \textbf{72.77\%}  \\ 
 \rowcolor{lightgray!30} STD (w/o bias) & 50\% & \textbf{55.41\%} & 56.91\% & \textbf{62.19\%} & 68.35\% & 71.82\%  \\ 
  \midrule 
Magnitude & 2:4 & 48.78\% & 52.09\% & 51.30\% & 51.78\% & 61.01\%  \\ 
 Wanda & 2:4 & \textbf{52.25\%} & \textbf{53.67\%} & \textbf{56.98\%} & 62.43\% & 68.27\%  \\ 
 \rowcolor{lightgray!30} STD (w/o bias) & 2:4 & 50.12\% & 52.33\% & 53.75\% & \textbf{63.85\%} & \textbf{68.75\%}  \\ 
  \midrule 
Magnitude & 4:8 & 49.80\% & 50.20\% & 51.38\% & 51.93\% & 64.33\%  \\ 
 Wanda & 4:8 & \textbf{52.80\%} & \textbf{53.99\%} & \textbf{61.01\%} & \textbf{66.61\%} & \textbf{69.85\%}  \\ 
 \rowcolor{lightgray!30} STD (w/o bias) & 4:8 & 52.09\% & 53.75\% & 56.98\% & 65.43\% & 69.22\%  \\ 
\bottomrule 
 \end{tabular} 
 \caption{Zero-shot performance on Winogrande \cite{ai2:winogrande}.} 
 \label{tab:winogrande} 
 \end{table*}

  \begin{table*}[h] 
 \centering 
 \begin{tabular}{ccccccc} 
 \toprule 
 Method & Sparsity & Qwen3-0.6B & Qwen3-1.7B & Qwen3-4B & Qwen3-8B & Qwen3-14B \\ 
 \midrule 
 Dense & 0\% & 56.20\% & 60.93\% & 66.06\% & 67.72\% & 72.85\% \\ 
 \midrule 
 Magnitude & 50\% & 49.88\% & 51.62\% & 51.62\% & 55.09\% & 64.80\%  \\ 
 Wanda & 50\% & 54.06\% & \textbf{57.38\%} & 62.12\% & \textbf{69.61\%} & \textbf{72.77\%}  \\ 
 \rowcolor{lightgray!30} STD (w/o bias) & 50\% & \textbf{55.41\%} & 56.91\% & \textbf{62.19\%} & 68.35\% & 71.82\%  \\ 
  \midrule 
Magnitude & 2:4 & 48.78\% & 52.09\% & 51.30\% & 51.78\% & 61.01\%  \\ 
 Wanda & 2:4 & \textbf{52.25\%} & \textbf{53.67\% }& \textbf{56.98\%} & 62.43\% & 68.27\%  \\ 
 \rowcolor{lightgray!30} STD (w/o bias) & 2:4 & 50.12\% & 52.33\% & 53.75\% & \textbf{63.85\%} & \textbf{68.75\%}  \\ 
  \midrule 
Magnitude & 4:8 & 49.80\% & 50.20\% & 51.38\% & 51.93\% & 64.33\%  \\ 
 Wanda & 4:8 & \textbf{52.80\%} & \textbf{53.99\%} & \textbf{61.01\%} & \textbf{66.61\%} & \textbf{69.85\%}  \\ 
 \rowcolor{lightgray!30} STD (w/o bias) & 4:8 & 52.09\% & 53.75\% & 56.98\% & 65.43\% & 69.22\%  \\ 
\bottomrule 
 \end{tabular} 
 \caption{Zero-shot performance on MMLU \cite{mmlu}.} 
 \label{tab:mmlu} 
 \end{table*}

\end{document}